# ROS Based Visual Programming Tool for Mobile Robot Education and Applications

Mustafa Karaca[1]

mustafa.karaca@inovasyonmuhendislik.com

Dr. Ugur Yayan[1]

ugur.yayan@inovasyonmuhendislik.com

November 27, 2020

## ABSTRACT

Visual programming languages (VPLs) provide coding without typing texts. VPL makes coding easy to programmers with automatically adding usually used some code structure. Beginners in coding have generally two main challenges; transforming ideas into logical expressions and syntax errors. Syntax errors are impossible with VPLs because of there is no forgotten parentheses and semicolons. VPLs provide to focus on algorithm for programmers. VPL is a new trend for educational robotic environments. In this study, Robot Operating System (ROS) compatible web based visual programming system has been developed for evarobot. ROS provides libraries and tools to help software developers create robot applications. It provides hardware abstraction, device drivers, libraries, visualizers, message-passing, package management, and more. Blockly has been used as VPL for the study and to generate / use blocks (commucation, sensing etc.). Some applications were generated like teleoperation, SLAM and wander etc. In this system, communication between server and client is supported by rosbridge package. Web page connected to ROS which runs on server using roslibjs library. Rosbridge provides a JSON API to ROS functionality for non-ROS programs.

*K*eywords Visual programming language, robot operating system, mobile robot, robotics, education, blockly

## 1. Introduction

People have to learn several information in the field of robotics because of growing technological productivity and developments in this area. We observe robotics works increasing in the market from day to day. Robots can do jobs like a human or they can do more. Nowadays robots have been playing an active role in many sectors therefore, robotic works are increasing day by day because of there is a need for qualified personnel in these sectors. Very few people know about robot programming, algorithm logic and follow projects on the subject. Learning programming from books or articles is very difficult and challenging. It can be done easier by visual perception. Therefore, robotic-based training becomes important. Robotics education and work in schools with supervisorship of assistant teacher is very interesting and extremely effective learning method for students. Visual Programming Languages (VPLs) are very good options for people who have no experience with text-based programming.

A visual programming language is any programming language that allow users create programs by manipulating program elements graphically rather than by specifying them textually and VPL is a new trend for educational robotic environments [1]. A VPL allows programming with visual expressions, spatial arrangements of text and graphic symbols used either as elements of syntax or secondary notation. For example, many VPLs (known as dataflow or diagrammatic programming) [2] are based on the idea of "boxes and arrows", where boxes or other screen objects are treated as entities, connected by arrows, lines or arcs which represent relations. Visual programming languages provide coding without typing texts. VPL makes coding easy to programmers with automatically adding usually used some code structure. Beginners in coding have generally two main challenges; transforming ideas into logical expressions and syntax errors. Syntax errors are impossible with VPLs because of there is no forgotten parentheses and semicolons. VPLs provide to focus on algorithm for programmers.

[1] Research and Development Department, Inovasyon Muhendislik Ltd. Sti., Eskisehir, Turkey

## 2. Related Works

The study [3], [4] and [12] emphasize the positive effects of robotics training on children. In study [3], a children club was founded at the Department of Computer Engineering of Joensuu University, in Finland. Robotic works has been performed with children whose ages ranging from 8 to 18, teachers and assistants. In the study it is aimed to develop technological solutions to individual needs of children and impact of applied technologies on learning and teaching. LEGO Mindstorms Robotics Invention System 2.0 and ELEKIT building sets were used in the study for creating a product. LEGO RIS 2.0 were used for programming LEGO robots and similarly TileDesigner for ELEKIT robots. In study [4], 11- and 12-years old children educated in "How should a robot designed" and have worked on this issue with partners from School of Engineering of University of Minho (SEUM). The study is performed at the Castêlo da Maia Basic School (CMBS), in Portugal. Teachers, assistants and parents have been supporting the children. After a year, students have achieved successful result in robot design competition RoboParty®. In study [12], a robot control and routing environment is formed using two computers and RoboEduc software at an elementary school in Brazil. The children see simple robotics parts and learn about them. Simple user interface is created to navigate robots for children. The application teaches robot programming to perform specific tasks, applying applications from simple level to higher levels. End of the study, self-confidence of children and ability of problem-solving, team work, physics, mathematics are improved.

At study [13-21] several applications have been developed because of challenge of creating algorithms using lines of code and syntax errors in the code. Several studies have been performed to providing easy of programming and simple usage for children. In study [13], a remote learning laboratory has been developed at University of Deusto Bilboa in Spain. Gathering objects in a maze scenario has been performed using only direction keys on the keyboard or blocks of VPL Google Blockly. User collects objects and controls the robot using created blocks. MSE ArduBot kit is placed on the Pololu RRC04A body of the robot. Python programming language is used in programming the robot. Graphical User Interface, Google Blockly, HTML, CSS and Javascript technologies were utilized when creating the user interface. Users can connect WebLab-Deusto remote laboratory from their personal computer and improve programming ability with educational robotics game. In study [14], an application has been developed for programming without high level programming knowledge. The purpose of the application is developing robot programs easily using visual blocks without any specific programming language. To realize this purpose a mobile application is created which name is Hammer. The application is used Google App Inventor and Scratch technologies and developed for Android smartphones and tablets. Robot created using 3D canvas and xml files. In study [15] a visual programming platform has been developed to teach and learn programming. Name of this application is Scratch and its target group is children whose age between 8 and 16. It is a multi-purpose educational application includes; interactive stories, games, animations, musics. When creating user interface LogoBlocks, LEGO MindStorms, LogoMicroworlds, Etoys, AgentSheets and Alice2 is used as reference. Squeak programming language and environment is used when developing Scratch. Scratch application is an educational platform that contribute to the development of children. An application has been conducted by high school students in India using the Scratch interface to develop embedded Arduino project. Application developed easily using Scratch due to opinion of students. In studies [16] and [17] an educational system P-CUBE has been developed for primary and secondary students in Japan. In the studies functional cubes were created to teach logic of programming for children. Children constitute an algorithm using these cubes and mobile robot acts depending on the algorithm. In addition, children can play games on the computer using the cubes. RFID tags are used in the construction of the cube and IR sensors are available on the mobile robot. The system contains programming material, the mobile robot, computer and programming blocks. programming training, different experience and robotics knowledge is given to children with this application. In study [18], a programming tool is developed to learn programming logic easily for children and the tool is named Algorithmic Bricks. This system contains a robot and visual objects to drive the robot using programming logic. An algorithm is created using objects and robot moves based on this algorithm. Thus, the programming logic and processes are understood by children simplistic way. In study [19] a visual programming interface is created for people who has no experience with programming. The interface is named LSOFT and designed for programming Mindstorms robots. Mindstrom NXT is a programmable robot kit manufactured by the Lego company. It provides an easier way to establish an algorithm because of it does not require typing any code with the keyboard to create an algorithm, it supports drag and drop technique with the created tools to perform specific functions. This interface makes programming robots much easier. Scratch [20] is a project of the Lifelong Kindergarten Group at the MIT Media Lab. It is provided free of charge. With Scratch, you can program your own interactive stories, games, and animations. You can share your creations with others in the online community. Scratch helps young people learn to think creatively, reason systematically, work collaboratively and essential skills for life in the 21st century. Scratch is designed especially for ages 8 to 16, but is used by people of all ages. Scratch projects in a wide variety of settings, including homes, schools, museums, libraries, and community centers. There is a step-by-step guide available inside Scratch. Rover [21] is an educational game developed by Nasa. User write an algorithm to move a rover using visual commands including move forward, turn right, turn left. Algorithm is designed for reaching specific locations on the map. According to this algorithm rover is trying to reach destination point while overcoming the obstacles and controlling its fuel gage.

In this study, Robot Operating System (ROS) compatible web based visual programming system has been developed for evarobot [5]. ROS provides libraries and tools to help software developers create robot applications. It provides hardware abstraction, device drivers, libraries, visualizers, message-passing, package management, and more. Evarobot is fully ROS compatible, low cost, multi-functional, transformable mobile robot purposed for educational and research developed by Inovasyon Muhendislik. Blockly [6] has been used as VPL for the study and to generate / use blocks (commucation, sensing etc.).

The rest of this paper is organized as follows. The system components which are Blockly, Robot Operating System and evarobot are explained in Section III. Section IV describes proposed visual programming language architecture as basic and advanced. Conclusion and future works are given in the final section.

## 3. System Components

Robotic block set has been developed using Google Blockly for robot control operations. The data collected from the sensors on the robot can be gathered and instructions can be sent to the robot using these blocks. Evarobot is used in this study which is a fully ROS compatible robot which is built by Inovasyon Muhendislik [5]. Blocks were created specifically for operations that can be performed by using evarobot. This study is performed by using open source software like ROS, Blockly, and Gazebo is presented to users as open source. Web page connected to ROS which runs on server using roslibjs [10] library. Server and robot are connected to each other using ROS. Communication between server and web page is supported by rosbridge server tool [9].

Blockly [6] is a visual editor that allows users to write programs by adding blocks together. Developers can integrate the Blockly editor into their own web applications to create a great UI for novice users. In this study, robotic block set has been developed using Google Blockly for robot control operations. The data collected from the sensors on the robot can be gathered and instructions can be sent to the robot using these blocks. Robot block set is constructed for non-coding users to control robot and realize the experiments. Two types of block set are generated. One of them is classed as basic VPL. Users who have no prior experience with robotics can programme a ROS compatible robot using blocks by using basic VPL architecture. Another block set are named as advanced VPL. Users who have experience with robotics and ROS can programmed robot using this set of blocks. These blocks provide more flexibility and advanced controls on robot to the user.

Gazebo is used to simulate Evarobot. Robot simulation is an essential tool in every roboticist's toolbox. A well-designed simulator makes it possible to rapidly test algorithms, design robots, and perform regression testing using realistic scenarios. Gazebo offers the ability to accurately and efficiently simulate populations of robots in complex indoor and outdoor environments. At your fingertips is a robust physics engine, high-quality graphics, and convenient programmatic and graphical interfaces. Best of all, Gazebo is free with a vibrant community. There is a package for using ROS and Gazebo together. This gazebo_ros_pkgs is a set of ROS packages that provide the necessary interfaces to simulate a robot in the Gazebo 3D rigid body simulator for robots.

Evarobot is fully compatible with ROS. Robot Operating System (ROS) provides libraries and tools to help software developers create robot applications. It provides hardware abstraction, device drivers, libraries, visualizers, message-passing, package management, and more. ROS is licensed under an open source, BSD license. The evarobot software system is written entirely in ROS and it is published at [8]. ROS is an open-source, meta-operating system for your robot. It provides the services you would expect from an operating system, including hardware abstraction, low-level device control, implementation of commonly-used functionality, message-passing between processes, and package management. It also provides tools and libraries for obtaining, building, writing, and running code across multiple computers.

In this system, communication between server and client (web page) is supported by rosbridge [9] package. Rosbridge provides a JSON API to ROS functionality for non-ROS programs. There are a variety of front ends that interface with rosbridge, including a WebSocket server for web browsers to interact with. roslibjs is the core JavaScript library for interacting with ROS from the browser. It uses WebSockets to connect with rosbridge and provides publishing, subscribing, service calls and other essential ROS functionality. Web page connected to ROS which runs on server using roslibjs [10] library. roslibjs is the core JavaScript library for interacting with ROS from the browser. It uses WebSockets to connect with rosbridge.

Evarobot [5] is fully ROS compatible, low cost, multi-functional, transformable mobile robot developed for educational and research purposed by Inovasyon Muhendislik. Evarobot consists of three main components modular mechanics, raspberry pi 2 compatible electronics and open source software (ROS) parts. The evarobot mechanical module consists of two parts body and the upper platform. Body contains motors, encoders, wheels, battery components and basic level sensors like infrared and sonar. Upper platform contains advanced sensors like Microsoft Kinect, 3600 RPlidar, and head angle reference system (MinIMU v9). Evarobot is a fully ROS compatible education and research purposed mobile robot for everyone who works on robotics. Gazebosim model and plugins of Evarobot have been created for the realization of the simulation environment for testing.

# 4. Proposed Visual Programming Language Architecture

Two types of block set are generated. One of them is classed as basic VPL. Users who have no prior experience with robotics can program a ROS compatible robot using blocks by using basic VPL architecture. Another block set are named as advanced VPL. Users who have experience with robotics and ROS can program robot using this set of blocks. These blocks provide more flexibility and advanced controls on robot to the user.

## 4.1 Basic Visual Programming Language Architecture

These blocks are generated for primary school, secondary school and high school students. They are simple blocks which are contains only basic movements and reading sensory data from sensors by using only one block.

- **i. Basic Movements:** Blocks under this category performs basic movements of robot. There are four basic movements includes move forward, move backward, turn left and turn right. Blocks under this category are shown in Fig. 1.

    Operations of these blocks are described from top to bottom.
    - Move Forward: This block moves robot forward to input parameter meter.
    - Move Backward: This block moves robot backward to input parameter meter.
    - Turn Left: This block rotates robot left to input parameter degree.
    - Turn Right: This block rotates robot right to input parameter degree.

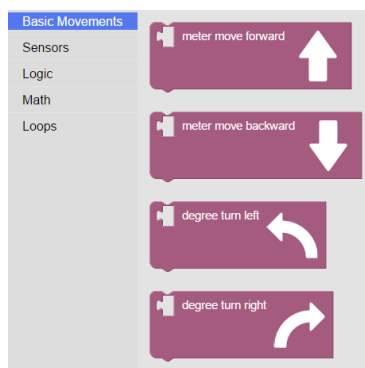

Figure 1: System (OTA) Configuration

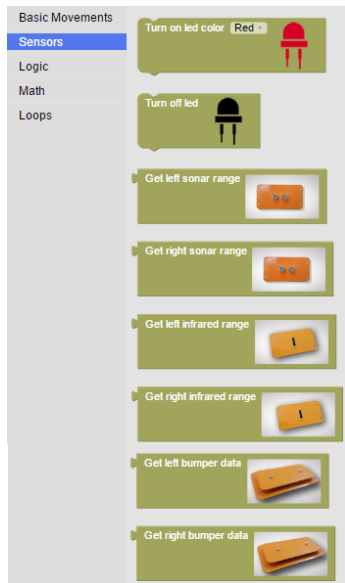

Figure 12: System (OTA) Configuration

ii. **Sensors:** There are four type of sensor control blocks under this category. These are RGB led, sonar, infrared and bumper. Blocks under this category are shown in Fig. 2.

　　Operations of these blocks are described from top to bottom

- Turn on led color: This block turns on led color as input parameter.
- Turn off led: This block turns off led.
- Get left sonar range: Returns range value of left sonar.
- Get right sonar range: Returns range value of right sonar.
- Get left infrared range: Returns range value of left infrared.
- Get right infrared range: Returns range value of right infrared.
- Get left bumper data: Returns data from left bumper.
- Get right bumper data: Returns data from right bumper.

iii. **Examples:** Fig. 3 shows code for drawing a square with robot movements. Firstly, robot moves 1 meter forward and then turns 90 degrees left. Robot repeats these operations 4 times.

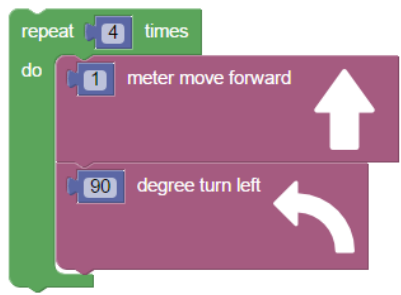

Figure 3: Drawing Square Example

**4.2 Advanced Visual Programming Language Architecture**

The connection between the server and evarobot was performed by using ROS. These two parts of the system provides controlling and driving the robot. The data collected from the sensors on the robot can be taken and instructions can be sent to the robot using the created ROS relevant blocks. Created blocks are categorized into three groups as ROS-Communication, ROS-Sensors and ROS-Behaviours.

**i. ROS-Communication:** This group includes communication-based ROS methods. Blocks under this category are shown in Fig. 4. Operations of these blocks are described from top to bottom.
- CONNECT TO ROS: It provides connecting to the server,
- CREATE PUBLISHER: It creates a data publisher to send data. Name of message and message type are parameters of the block,
- PUBLISH DATA: This block is used to send data over created publisher. Data is input of the block,
- CREATE SUBSCRIBER: It creates a subscriber to receive data. Name of message and message type are parameters of the block,
- SUBSCRIBE: Operations to be performed are determined in this block when data received from determined subscriber.

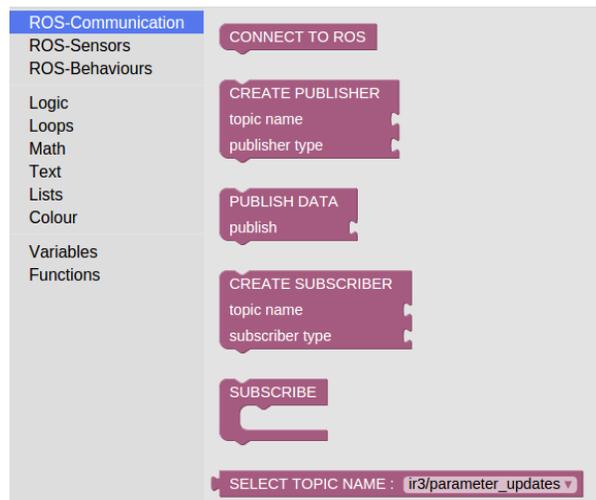

Figure 4: ROS-Communication Blocks

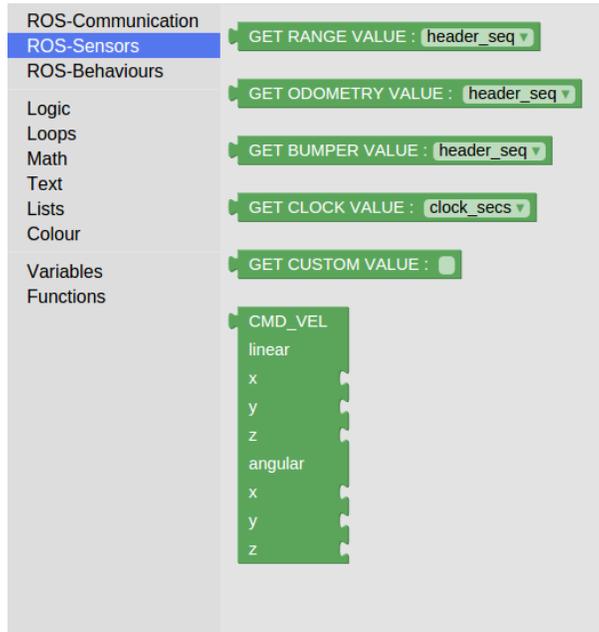

Figure 5: ROS-Sensor Blocks

- **ii. ROS-Sensors:** This group includes sensor-based ROS methods. Blocks under this category are shown in Fig. 5. Operations of these blocks are described from top to bottom.
    - CONNECT TO ROS: It provides connecting to the server,
    - CREATE PUBLISHER: It creates a data publisher to send data. Name of message and message type are parameters of the block,
    - GET RANGE VALUE: Returns selected value of created subscriber. Message type of the subscriber should be sensor_msgs/Range,
    - GET ODOMETRY VALUE: Returns selected value of created subscriber. Message type of the subscriber should be nav_msgs/Odometry,
    - GET BUMPER VALUE: Returns selected value of created subscriber. Message type of the subscriber should be impc_msgs/Bumper,
    - GET CLOCK VALUE: Returns selected value of created subscriber. Message type of the subscriber should be rosgraph_msgs/Clock,
    - GET CUSTOM VALUE: Returns typed value of created subscriber,
    - CMD_VEL: Returns cmd_vel values for a geometry_msgs/Twist type publisher.

- **iii. ROS-Behaviours:** This group includes sensor-based ROS methods. Blocks under this category are shown in Fig. 5. Operations of these blocks are described from top to bottom.
    - TELEOP: This block listens arrow keys and navigates the robot due to the pressed key. Up arrow key for moving forward the robot, down arrow key for moving back the robot, right arrow key for turning robot right, left arrow key for turning robot left and spacebar key for stopping the robot,
    - WANDER: In this mode the robot autonomously wanders around avoiding obstacles using sensor readings. If sensor range value lower than threshold value robot turns and goes ahead,

- SET PID PARAMETERS: This block set pid parameters for left and right wheel. Takes kp, kd, ki values as input for both wheel

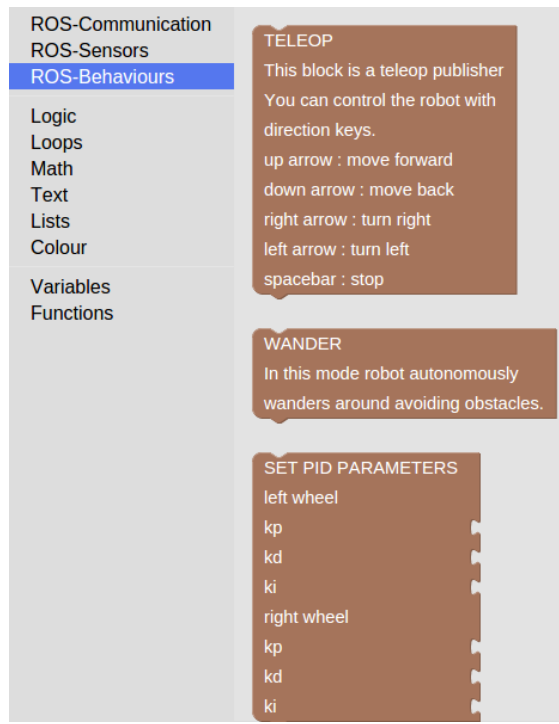

Figure 6: ROS-Behaviour Blocks

**iv.** **ROS-Examples:** : In this section there are number of activities to show usage of the above blocks. Fig. 7 show how to subscribe to a sonar data publisher. Firstly, connect to ROS. Then, create a subscriber with parameter topic name and message type. Then, specify operations that is going to execute when data received. In this example range value of received data is alerted. field_of_view, min_range, max_range and range options are available in sonar data.

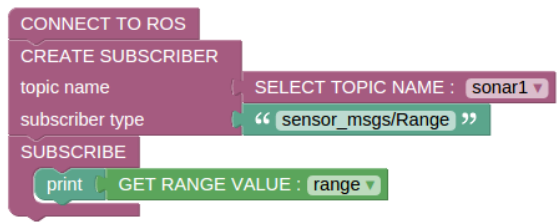

Figure 7 Subscribe to Sonar

Fig. 8 shows publishing data to cmd_vel topic. Firstly, connect to ROS. Then, create a publisher with parameter topic name and message type. Then, send data. In this example cmd_vel value is sent. This data takes linear and angular velocities in x, y, z directions.

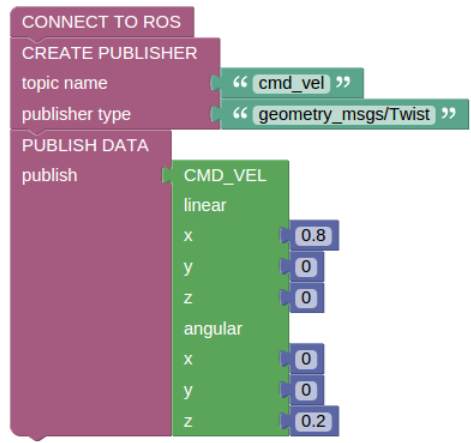

Figure 8: Publish cmd_vel

In this mode the robot autonomously wanders around avoiding obstacles on its own and sends the sensor readings. Some part of wander code and gazebo image of evarobot is shown in Fig. 9. Wander code is written by using prepared special blocks.

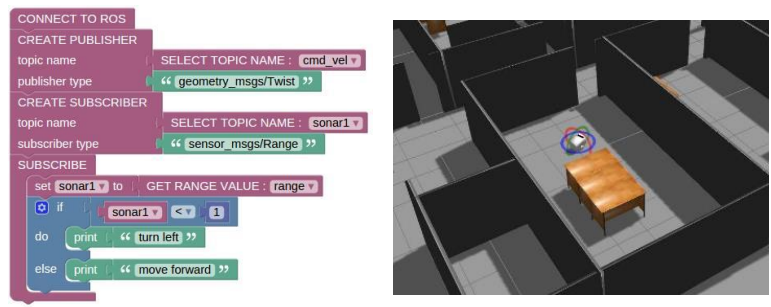

Figure 9: Evarobot Wander Code

## 5. Conclusion

In this study, ROS compatible web based visual programming system has been developed for EvaRobot. Advanced visual programming language architecture is explained. Blockly has been used to generate blocks and use them. Some applications were generated like subscriber, publisher and wander mode. In the future, system is planned to be converted into a visual programming based remote access laboratory.